\documentclass[letterpaper, 10 pt, conference]{ieeeconf}  % Comment this line out if you need a4paper
\usepackage{import}
\usepackage{subcaption}
\usepackage{threeparttable}
\usepackage{amsmath}
\usepackage{amssymb}  % assumes amsmath package installed
\usepackage{bm}
\usepackage{booktabs}
\usepackage{makecell}
\usepackage{multirow}
\usepackage[table]{xcolor}
\usepackage{balance}
\usepackage{csquotes}
\usepackage{listings}
\lstdefinestyle{custompython}{
  language=Python,                     % the language of the code
  basicstyle=\ttfamily\small,          % the size and typeface of the font
  keywordstyle=\color{blue},           % color for keywords like "for", "if", "while", etc.
  identifierstyle=\color{black},       % color for identifiers (e.g., variable names)
  commentstyle=\color{green},          % color for comments
  stringstyle=\color{red},             % color for strings
  showstringspaces=false,              % don't show spaces in strings
  columns=flexible,                    % make columns flexible to avoid character overlapping
}
\usepackage{todonotes}
\IEEEoverridecommandlockouts                              % This command is only needed if 
\title{\LARGE \bf
Bridging the Reality Gap of Reinforcement Learning based Traffic Signal Control using Domain Randomization and Meta Learning 
}

\author{Arthur Müller$^{1}$ and Matthia Sabatelli$^{2}$% <-this % stops a space
\thanks{*The work was partly supported by the Ministry of Economic Affairs, Industry, Climate Action and Energy of the State of North Rhine-Westphalia, Germany, under the project ”SUPPORT” (005-2111-0026).}% <-this % stops a space
\thanks{$^{1}$A. Müller is with the Fraunhofer IOSB-INA, 32657 Lemgo, Germany:
        {\tt\footnotesize arthur.mueller@iosb-ina.fraunhofer.de}}
\thanks{$^{2}$M. Sabatelli is with the Department of Artificial Intelligence and Cognitive Engineering, University of Groningen, 9712 CP Groningen, The Netherlands:
        {\tt\footnotesize m.sabatelli@rug.nl}}% \small
}

\begin{document}

\maketitle
\thispagestyle{empty}
\pagestyle{empty}

%%%%%%%%%%%%%%%%%%%%%%%%%%%%%%%%%%%%%%%%%%%%%%%%%%%%%%%%%%%%%%%%%%%%%%%%%%%%%%%%
\begin{abstract} %-max ? words
% Setup: general context on the broader picture. Should be comprehensible to a scientist in any discipline. Should describe the general interest in the topic (few sentences)
Reinforcement Learning (RL) has been widely explored in Traffic Signal Control (TSC) applications, however, still no such system has been deployed in practice. A key barrier to progress in this area is the \textit{reality gap}, the discrepancy that results from differences between simulation models and their real-world equivalents. In this paper, we address this challenge by first presenting a comprehensive analysis of potential simulation parameters that contribute to this reality gap. We then also examine two promising strategies that can bridge this gap: Domain Randomization (DR) and Model-Agnostic Meta-Learning (MAML). 
% Beide wurden mit einem Verkehrssimulationsmodell einer Kreuzung trainiert. Zudem wurde das Modell in LemgoRL eingebettet, einem Framework, das realistische, sicherheitskritische Anforderungen in die Steuerung integriert. Die Evaluierung wurde mit einem anderen Modell derselben Kreuzung durchgeführt, welche mit einem anderen Verkehrssimulator erstellt wurde.
Both strategies were trained with a traffic simulation model of an intersection. In addition, the model was embedded in LemgoRL, a framework that integrates realistic, safety-critical requirements into the control system. 
Subsequently, we evaluated the performance of the two methods on a separate model of the same intersection that was developed with a different traffic simulator. In this way, we mimic the reality gap.
% They are trained and evaluated in two different microscopic traffic simulators, SUMO and VISSIM. 
% Both models are also embedded in LemgoRL, a framework that incorporates realistic safety requirements. 
Our experimental results show that both DR and MAML outperform a state-of-the-art RL algorithm, therefore highlighting their potential to mitigate the reality gap in RL-based TSC systems.

\end{abstract}
%%%%%%%%%%%%%%%%%%%%%%%%%%%%%%%%%%%%%%%%%%%%%%%%%%%%%%%%%%%%%%%%%%%%%%%%%%%%%%%%

\section{Introduction}
% 	- Hourglass shape like the abstract, but in contrast to the abstract, the hourglass of the introduction should be top-heavy, emphasising the context - the top of the hourglass - more than the resolution of the story
% 	- The first sentences are the most important ones ("lede"). They decide if a reader stays or stays not. Make them interesting and strong
% 	- 4 paragraph approach
% 		1. Broad context and background for your research. Introducing problem area and knowledge gap. Giving a short overview of the state of the art, explaining what has already been discovered. You can make your first paragraph stronger by ending it with some contrast or cliffhanger.
Traffic Signal Control (TSC) plays a crucial role in managing traffic flow and reducing congestion in urban areas. Reinforcement learning (RL), a machine learning technique, has been intensively investigated as a promising approach to address the shortcomings of existing TSC strategies. But up to date, rarely have RL applications made their way to real-world deployment \cite{Dulac-Arnold2021}. RL-based TSC has even never been deployed in the real world yet \cite{Chen2022}\cite{Wei2021}. 

% - traffic signal controller -> goal of better traffic control than with conventional control strategies -> a lot of potential for improvement
% - RL is a promising approach to overcome the shortcomings of current tsc 
% - RL is machine learning technique
% % - initiated thru the success of Deepmind in 2013 -> a lot of successes in different areas
% - a lot of research has been done in the field of RL-based TSC (many paper publications)
% - But RL in general has rarely made its way to reality due to several challenges \cite{Dulac-Arnold2021}
% - This is particularly true for RL based tsc: It has not made the way to real world deployment yet due to several reasons \cite{Chen2022}\cite{Wei2021} 
% - \cite{Chen2022} gives an excellent overview about the challenges that are to overcome to bridge the gap to real-world deployment: Uncertainty in detection, Reliability of communications, Compliance and interpretability, Heterogeneous road users

% 		2. This paragraph switches to a close-up point of view and zooms in on your particular research problem. Broader motivation to the exact question. At the end of this paragraph, you should clearly and explicitly state the research question that your paper addresses.
The primary reason for this is the reliance on simulation for RL training in TSC. Simulation offers various benefits, such as faster training, scalability, cost-effectiveness, and safety. Safety is a crucial requirement since TSC is a highly safety-critical domain. Failures could harm people or vehicles. However, transferring learned behaviors from simulated environments to the real world is challenging \cite{Tobin2017} due to differences between the two. 
These differences are known as the \textit{simulation-to-reality gap} \cite{Zhao2020} or just \textit{reality gap} \cite{Dimitropoulos2022}\cite{Tobin2017}. 
They can manifest, for example, in the geometry of intersections, the behavior of the drivers, or noise in traffic detection sensors.
One contribution of this paper is identifying and providing a comprehensive list of potentially relevant simulation parameters that constitute the reality gap.
% \cite{Chen2022} gives a broad overview about the barriers to applying RL-based TSC in real-world settings.

Various methods have been proposed to bridge the reality gap. They have been investigated especially in the field of robotics. Nonetheless, as noted by \cite{Chen2022}, TSC presents unique challenges that are not typically encountered in robotics. It is, therefore, not clear, if successful methods in the field of robotics are also successful in the field of TSC. Consequently, our paper aims to investigate the effectiveness of two important representatives of these methods, namely Domain Randomization (DR) and Meta Reinforcement Learning (Meta-RL), in overcoming the reality gap in the domain of TSC. As the base algorithm for both methods, we use a Proximal Policy Optimization (PPO) algorithm \cite{Schulman2017}. We measure the effectiveness by training and evaluating the algorithms in two different simulation models of the same intersection created with two different simulation tools. This emulates the reality gap.

As a basis for our investigation, we choose a framework for training and evaluating that is most advanced in respect to the deployability in reality. To the best of our knowledge, this is the LemgoRL framework \cite{LemgoRL2022}\cite{Mueller2022}. It was highlighted from a recent survey paper \cite{Chen2022} as leading the way for further development towards reality deployment, since it "incorporates realistic domain constraints" \cite{Chen2022} and "RL algorithms that are trained using such benchmarks would likely have better generalizability and robustness in deployments" \cite{Chen2022}.

Our experiments show that DR and Meta-RL significantly outperform a PPO agent trained without any means of simulation-to-reality adaption. As such, our research demonstrates the applicability and efficacy of both DR and Meta-RL as potent techniques to bridge the reality gap in RL-based TSC systems.

\section{Reinforcement Learning \\ for Traffic Signal Control}
\label{sec:RL}
\subsection{Reinforcement Learning}
In Reinforcement Learning an agent interacts with its environment to achieve an optimization goal. At each time step $t$, the agent takes an \textit{action} $a_t$ and receives an observation from the environment which constitutes the \textit{state} $s_t$ of the environment. Additionally, the agent receives a \textit{reward} $r_{t+1}$ at each time step 
defined by the reward function $R(s_t, a_t)$ 
indicating how good or bad the immediate situation is in terms of the optimization goal. The decision of which action is taken at each time step is made by the policy $\pi(a_t|s_t)$, which maps states to actions. The agent tries to collect as much reward as possible in the long run, which formally corresponds to finding an optimal policy $\pi^*$ that maximizes the expected cumulative reward over time
\begin{equation}
\pi^* = argmax_{\pi} E[\Sigma_{t=0}^T \gamma^t * R(s_t, a_t) | \pi],
\end{equation}
where T is the time horizon and $\gamma \in [0,1)$ is the discount factor. 

% $$V_{\pi}(s) = E[\Sigma_{t=0}^T \gamma^t * R(s_t, a_t)]$$

It is well known that RL problems are typically formulated as Markov Decision Processes (MDPs), defined as a 5-tuple \cite{Sutton2018} consisting of a set of states $\mathcal{S}$, a set of actions $\mathcal{A}$, a transition probability function $\mathcal{T}(s_{t+1} | s_{t}, a_{t})$, a reward function $R(s_t, a_t)$, and a discount factor $\gamma$.

\subsection{Traffic Signal Control as RL problem}
% - RL-based TSC is widely studied \cite{Noaeen2022}\cite{Haydari2020}: viele Publikationen, viele Methoden vorgestellt
% - aber wie schon erwähnt, kein real world deployment so far \cite{Chen2022}
% - da unsere Arbeit darauf abzielt, RL-based TSC in der Realität verfügbar zu machen, macht es Sinn, ein Framework zu nehmen, dass am weitesten hinsichtlich seiner Einsatzfähigkeit in der Realität ist
% - Given these considerations, we have selected the LemgoRL framework for conducting our experiments.
% - Denn LemgoRL integriert realisische Constraints, die sicherheitsrelevante Anforderungen befriedigen
% - LemgoRL "mimic the complexities and intricacies", safety constraints. Ein RL-Agent in diesem Framework wurde sogar in der Echtwelt eingesetzt

The realm of RL-based Traffic Signal Control is vast and has been the subject of comprehensive studies, as demonstrated by numerous publications presenting a diverse range of methods \cite{Noaeen2022}\cite{Haydari2020}. However, as previously mentioned, none of these methods have been deployed in real-world scenarios \cite{Chen2022}.

Given our objective of bridging the reality gap, we choose the LemgoRL framework that is most advanced in terms of its potential for real-world deployment.
% Given our objective of bridging the reality gap, it is logical to choose a framework that is most advanced in respect to potential real-world deployment. Therefore, we choose the LemgoRL framework.
% It provides a simulation model of an intersection in Lemgo, Germany, for training RL algorithms. 
% It features a realistic model of the road network made with SUMO, an open-source microscopic traffic simulator \cite{Lopez2018}. 
It provides a realistic model of an intersection in Lemgo, Germany, with 4 streets and 4 crosswalks. LemgoRL was made with SUMO, an open-source microscopic traffic simulator \cite{Lopez2018}.  
Furthermore, all relevant road user groups (vehicles, bicyclists, and pedestrians) as well as different vehicle classes (passenger cars, motorcyclists, trucks, trucks with trailers, and buses) and a realistic traffic volume during rush hour are included in the framework.  

% - \cite{Chen2022}: "Heterogeneous road users.(Section 7) Most simulations for RL-based TSC assume that all cars are the same size and have the same free-flow speed. However, cars share the road with pedestrians, buses, and emergency vehicles. Algorithms must detect and respond to the needs of different road users in a safe, equitable manner" --> For chen considering heterogneous road users crucial for real-world deployme nt --> LemgoRL does this. Includes heterogenous road users in simulation model and also pedestrians in reward function

Most importantly, LemgoRL ensures compliance with all relevant safety requirements, so that a real-world deployment of an RL agent would be permissible. On the one hand, this is realized by a safe-by-design action space. The action space consists of 8 traffic phases, where each phase represents a combination of non- or partially conflicting traffic flows. Collision of vehicles because of a simultaneous green signal for conflicting traffic flows is therefore not possible. 
% The relationship of the traffic phases to each other are shown in Fig. \ref{fig:traffic_phases}) where phases are represented as nodes and possible phase transitions as edges. Phase 2 for example allows vehicles to move on the west-east-axis and pedestrians to cross the street in the north and south. From phase 2 it is only allowed to make a transition to phase 3 (only vehicles on the west-east-axis).      

% \begin{figure}
%     \centering
%     \includegraphics[width=0.48\textwidth]{graphics/Phasendiagramm.pdf}
%     \caption{Action Space Traffic Phases \cite{Müller2022}.}
%     \label{fig:traffic_phases}
% \end{figure}

On the other hand, compliance with all safety regulations is ensured by the use of a Traffic Signal Logic Unit (TSLU). The TSLU has been implemented by traffic engineers using the professional traffic engineering software LISA\footnote{https://www.schlothauer.de/en/software-systems/lisa/}. It incorporates all rules that are relevant to traffic law. When RL agents compute an action, it is first checked by the TSLU to determine whether the phase request is allowed or not. If the phase request is not allowed, the system remains in the current phase or moves to the next allowed phase. In addition, the TSLU ensures that minimum and maximum phase duration are met. Furthermore, it guarantees compliance with all intermediate times, such as transition and interval times, that must be adhered to during certain phase transitions. These times are important to ensure sufficient time for road users to safely cross the road or intersection before a conflicting traffic flow would get a green signal. 
% For the architecture of LemgoRL see Fig.\ref{fig:LemgoRL}.

% Another means by achieving the necessary level of safety and readiness for deployment is by using a \textit{traffic signal logic unit} (TSLU). The TSLU is implemented with the professional traffic engineering software LISA and incorporates all relevant rules, which guarantees compliance with all safety and regulatory requirements at all times. When the RL agent computes the next action, representing the desired next traffic phase, 

% \begin{figure}
%     \centering
%     \includegraphics[width=0.48\textwidth]{graphics/Process_Diagram.pdf}
%     \caption{Architecture of LemgoRL\cite{LemgoRL2022}.}
%     \label{fig:LemgoRL}
% \end{figure}
Regarding the definition of the state space and the reward function, we follow [9] to a great extent.
The state space combines information about all incoming lanes regarding the number of vehicles (\texttt{wave}), the number of queued vehicles (\texttt{queue}), and the maximum waiting time of a vehicle (\texttt{wait\_veh}) in a vector. 
These are traffic information that can be detected with camera or radar sensor systems. In addition, the state space encompasses \texttt{wait\_ped}, which represents the time a pedestrian is waiting at a crosswalk for a green signal. 
The reward function targets minimizing the number of queued vehicles and the waiting time for vehicles and pedestrians and is defined as:
% \begin{multline*}
%     r_{t+1} = - \sum_{l} \Bigl(\alpha_{q} \cdot \texttt{queue}_{t+1}[l]  +  \alpha_{w,veh} \cdot \\  \texttt{wait\_veh}_{t+1}[l] + \alpha_{w,ped}\cdot \texttt{wait\_ped}_{t+1}[l]\Bigr),
% \end{multline*} 
%  where $l$ denotes the specific incoming lane or crosswalk of the intersection. The coefficients $\alpha_{q}$, $\alpha_{w,veh}$, and $\alpha_{w,ped}$ balance the optimization goals.  

\begin{multline}
    r_{t} = - \sum_{l_{p}} \Bigl( \alpha_{w,ped}\cdot \texttt{wait\_ped}_{t}[l_{p}]\Bigr) 
    \\- \sum_{l_{v}} \Bigl(\alpha_{q} \cdot \texttt{queue}_{t}[l_{v}]  +  \alpha_{w,veh} \cdot  \texttt{wait\_veh}_{t}[l_{v}]\Bigr),
\end{multline} 

 where $l_{v}$ denotes an incoming lane and $l_{p}$ a crosswalk of the intersection. The coefficients $\alpha_{q}$, $\alpha_{w,veh}$, and $\alpha_{w,ped}$ balance the optimization goals.

\section{The Reality Gap in TSC Simulation}
\label{sec:RealityGap}
The problem of the reality gap in the context of RL is mainly studied in the field of robotics \cite{Muratore2021}\cite{Dimitropoulos2022}\cite{Arndt2020}\cite{Andrychowicz2020}\cite{Dai2019}\cite{Zhao2020}\cite{Peng2018}. 
% where the goal is to learn robotics skills in simulation and transfer these skills to reality. 
As of now, there is no comprehensive study about the relevant differences between simulation and reality when it comes to transferring learned policies in the TSC domain. Therefore, we take a diverse set of different aspects into account so as not to miss out on the crucial ones. In a traffic simulation, there are several areas, where a discrepancy to reality could have an impact when transferring a TSC policy from simulation to the real world: \textit{driving behavior}, \textit{vehicle properties}, \textit{traffic volume}, \textit{intersection geometry}, and \textit{sensor noise}. As we are using SUMO for the training phase of our experiment, we show how these areas can be modeled and parameterized in this simulation tool.   

\subsection{Driving Behavior and Vehicle Properties}
In microscopic traffic simulation, vehicle movements and interactions within a road network are modeled individually. They are mainly determined by the physical properties of the vehicle and the driving behavior. These properties determine the underlying dynamics of the system, such as the average number of vehicles that can pass through an intersection during a green phase, and can potentially impact the transfer of TSC policies from simulation to reality. 
% Various parameters can be used to model driving behavior and vehicle properties.

\textbf{Car-following models} are at the core of modeling the driving behavior in microscopic traffic simulations. They describe how vehicles follow each other in a lane, i.e. how they react to changes in the leading vehicle's speed and distance. 
% Car-following models are a well established research field that brought about a variety of different model classes over the years like \textit{safety distance} models, \textit{optimal velocity} models, or \textit{psychophysical} models\cite{Li2012}\cite{Matcha2020}. 
For our experiments, we use a modification of the Krauss model \cite{Krauss1998} which is the default model in SUMO, and the Intelligent Driver Model (IDM) \cite{Treiber2000}, since both are widely used in the SUMO community \cite{BiekerWalz2017}\cite{Pourabdollah2017}.

The Krauss model lets vehicles drive as fast as the speed limit permits, while a vehicle is always able to decelerate fast enough to avoid a collision with the leading vehicle if it begins to brake. Therefore, a driver tries to keep a minimum time headway to the leading vehicle to react fast enough. This time headway is represented by \texttt{tau}. A smaller value of \texttt{tau} leads to a smaller desired time headway, which results in a more aggressive driving behavior since the distance to the leading vehicle is shorter. 
% Conversely, a larger value leads to longer distances. 
The parameter \texttt{sigma} models the driver's imperfection by randomly varying the acceleration of a vehicle, which represents different reaction times drivers have in reality.
% Another important parameter in the Krauss model is $\sigma$. It models the driver's imperfection by randomly varying the acceleration of a vehicle, which represents different reaction times drivers have in reality.

IDM is an alternative car-following model that determines the acceleration of a vehicle based on its current speed and the speed of the leading vehicle as well as the distance to the leading vehicle. An important parameter of this model is \texttt{delta}, the acceleration exponent. It determines the shape of the acceleration function, i.e. how rapidly a vehicle reacts to changes in speed or distance to the leading vehicle. A higher value of \texttt{delta} leads to more aggressive acceleration behavior.

Besides the car-following model, other parameters determine the driving behavior as well. For each individual vehicle, the road speed limit will be multiplied with a randomly sampled value for the parameter \texttt{speedFactor}, which reflects the vehicles' movements of reality more accurately. 
% resulting in the so-called \textit{free flow driving speed} of that vehicle. That is the vehicle's speed when no other vehicles or obstacles are in its way. 
% Randomizing \texttt{speedFactor} and therefore free flow driving speed of all vehicles reflects the vehicles movements of the reality more accurately than having the same free flow driving speed for all vehicles.
% Randomizing \texttt{speedFactor} for all vehicles reflects the vehicles' movements of reality more accurately.
Two other parameters in this context are \texttt{minGap} and \texttt{jmStoplineGap}. The former specifies the distance of a vehicle to the leading vehicle when there is a jam. The latter defines the distance of a halting vehicle to a traffic-light-controlled stop line. \texttt{impatience} is another parameter we take into account. It represents the willingness of drivers who are in a congestion situation or waiting at an intersection to impede other vehicles, forcing them to slow down.

Furthermore, the physical properties of the vehicle itself shape the dynamic of the system. The most important ones are \texttt{accel}, \texttt{decel}, and \texttt{length}. \texttt{accel} represents the acceleration ability of a vehicle while \texttt{decel} the deceleration ability. \texttt{length} corresponds to the modeled length a vehicle has in the simulation.  
    % - Randomize acceleration and decceleration of vehicles(accel, decel)
    %     % https://SUMO.dlr.de/docs/Definition_of_Vehicles%2C_Vehicle_Types%2C_and_Routes.html#vehicle_types
    %     "Please note that even though the car-following parameters are describing values such as max. acceleration, or max. deceleration, they mostly do not correspond to what one would assume. The maximum acceleration for example is not the car's maximum acceleration possibility but rather the maximum acceleration a driver choses - even if you have a Jaguar, you probably are not trying to go to 100km/h in 5s when driving through a city."
    % - Randomize vehicle lengths (length)
    %     "Due to the work on car following models, we decided to use two values for vehicle length. The length-attribute describes the length of the vehicle itself."

    % - longitudinal and lateral direction ??
    % - image like in \cite{Huang2021}

% \begin{figure}
%     \centering
%     \includegraphics[width=0.48\textwidth]{graphics/carfollowing_gapacceptance_langechanging.PNG}
%     \caption{Visualization of car-following, lane-change and gap-acceptance \cite{Huang2021}.}
%     \label{fig:dr_da_ida}
% \end{figure}

% \subsection{Vehicle Properties}
% Vehicle Properties
% - what is vehicle properties?
% - why it impacts TSC transfering
% - which parameters model it

\subsection{Traffic Volume}
Differences in traffic volume, or the number of vehicles crossing an intersection within a specific time interval, can also contribute to the reality gap. 
% These discrepancies may alter the underlying dynamics of the system. 
Traffic volume differences might arise when a simulation is based on outdated measurement data. Traffic volumes typically change over time due to local developments (e.g., the establishment or closure of businesses, schools, etc.). Additionally, differences may result from uncertainty in measurement methods, such as Bluetooth measurement campaigns, where the number of vehicles is estimated based on the presence of passing Bluetooth devices. A simple approach to scaling traffic volume in SUMO is given through \texttt{scale}. This parameter enables the generation of different traffic volumes during training.

% A simple approach to scaling traffic volume and adjusting the number of vehicles in a simulation is provided in SUMO through the \texttt{scale} parameter. This parameter enables the generation of different traffic volumes during training, which can help account for potential discrepancies between the simulation and real-world traffic conditions.

% - Das Verkehrsaufkommen, also die Anzahl an Fahrzeugen, die die Kreuzung in einem bestimmten Zeitintervall überqueren, kann sich zwischen Simulation und Realität unterscheiden. Dadurch ändert sich auch die zugrundeliegende Dynamik des Systems. 
% - Der Unterschied kann z.B. dadurch entstehen, dass die Simulation auf veralteten Messdaten beruht. Verkehrsaufkommen ändern sich typischerweise über die Zeit aufgrund von lokalen Veränderungen (Hinzukommen oder Wegfallen von Geschäften, Schulen etc.) oder anderen Faktoren. 
% - Der Unterschied kann aber auch durch unsicherheit-behafteten Messmethoden entstehen wie das bei Bluetooth-Messkampagnen der Fall ist, bei der die Anzahl der Fahrzeuge aufgrund von vorbeifahrenden Bluetooth-Geräten geschätzt wird.  

% - Eine einfache Möglichkeit das Verkehrsaufkommen zu skalieren und damit die Anzahl der Fahrzeuge in der Simulation anzupassen, wird bei SUMO durch den Parameter \texttt{scale} ermöglicht.
% - Damit lassen sich während des Trainings unterschiedliche Verkehrsaufkommen erzeugen.

\subsection{Intersection Geometry}
One aspect of the reality gap in TSC simulation stems from discrepancies in the geometric dimensions of roads and intersections. These discrepancies can arise when, for instance, SUMO models are created based on OpenStreetMap (OSM) \cite{OSM2023} data, which is a common practice. 
% OSM contains worldwide geographical information for free, which are collected and contributed by volunteers. 
The data from OSM can sometimes be outdated, have simplified geometries, or contain imprecise positional information. As a result, features such as stop lines in front of traffic signals might be inaccurately positioned, or the size of the intersection might differ between the simulation and the real world.

% The size of the intersection, for example, influences the number of vehicles that can occupy it at any given time. This becomes particularly crucial in situations involving left-turning traffic. Vehicles intending to turn left must often yield to oncoming traffic, causing them to enter the intersection and wait in its center. Any discrepancy between the simulated and real-world intersection sizes affects the number of vehicles that can wait in the center of the intersection.

\subsection{Sensors}
\label{subsec:sensors}
Sensor malfunctions and sensor noise pose significant challenges when transferring RL strategies to real-world settings in general \cite{Dulac-Arnold2021} and particularly in the TSC domain \cite{Chen2022}. Traffic sensing devices can be broadly classified into two groups: intrusive detectors, that are embedded within the road surface (e.g., induction loop detectors), and non-intrusive detectors, positioned above the road surface \cite{Sunkari2019}\cite{Tasgaonkar2020}.

\begin{figure}
    \centering
    \includegraphics[trim=0 1cm 0 1cm, clip, width=0.37\textwidth]{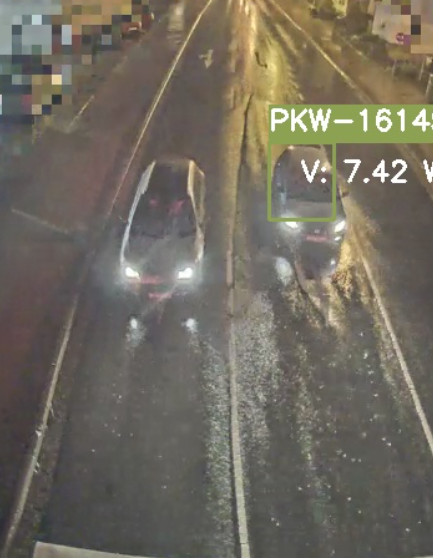}
    \caption{A snapshot of the camera sensor at the Lemgo intersection on a rainy night, showing that a vehicle was not detected due to poor visibility.}
    \label{fig:sensor_noise}
\end{figure}

Induction loop detectors are among the most frequently used traffic detectors. While they provide overall accurate measurements, they are also susceptible to deterioration over time \cite{Mills1998}. Lately, there has been an increasing shift towards non-intrusive detectors, such as video-based sensor technologies. However, these systems encounter diminished accuracy during adverse weather conditions (e.g., rain or snow) or at night \cite{Tasgaonkar2020}. Fig.~\ref{fig:sensor_noise} shows a snapshot captured by the camera sensor at the intersection in Lemgo during a night with rain. It shows how a vehicle is not detected due to poor visibility conditions.

Considering these constraints, coping with "uncertainty in detection" \cite{Chen2022} is recognized as a primary challenge when it comes to deploying RL agents as controllers in the real-world. 

\section{Methods to Bridge the Reality Gap}
\label{sec:methods}

% \begin{figure}
%     \centering
%     \includegraphics[width=0.48\textwidth]{graphics/sim-2-real-transfer_robotics.PNG}
%     \caption{Sim-2-Real-Transfer \cite{Zhao2020}.}
%     \label{fig:sim-2-real-transfer}
% \end{figure}

% \begin{figure}
%     \centering
%     \includegraphics[width=0.48\textwidth]{graphics/dr_and_domain-adaptation_indea.PNG}
%     \caption{Domain Randomization and Domain Adaption \cite{Zhao2020}.}
%     \label{fig:dr_da_ida}
% \end{figure}

Transferring policies that are trained in a simulation environment to the real world is a challenging task, since a simulation does not perfectly match all aspects of the real world. 
% The reality gap is a major challenge when it comes to transferring policies learned in simulation to reality. 
% This is because modeling errors or inaccuracies can cause policies that are optimal in simulation to be unsuccessful in reality. 
Errors or inaccuracies in a simulation model can cause policies that are optimal in simulation to be unsuccessful in reality.
This is because RL algorithms tend to exploit error-prone aspects of the model or learn policies that are infeasible in reality 
% if it brings benefits 
\cite{Muratore2021}\cite{Peng2018}. The algorithm would overfit to this faulty environment, which is why performance in reality decreases or the learned algorithm is entirely unusable. Therefore, much research has been done in recent years, especially in the robotics community, to develop methods that can overcome this reality gap.

One possibility is to make the simulation as realistic as possible. However, there is a consensus, at least in the robotics community, that this alone is not sufficient to fully close the gap, as perfect realism is not achievable \cite{Muratore2021}. Therefore, in addition to efforts to develop more realistic simulation models, methods have been developed to deal with imperfect simulation models. \cite{Dimitropoulos2022} categorizes these as (1) simulation-only methods and (2) adaptation methods. In the first category, an attempt is made to learn a robust behavior through sufficient variation in the simulation that considers the real world as a further variation. The main representative method in this category is Domain Randomization. The second category includes methods incorporating real-world data into the training process, such as Meta Reinforcement Learning. In this paper, we investigate the suitability of methods from both categories: Domain Randomization and Model-Agnostic Meta Learning (MAML) \cite{Finn2017}, which can be utilized as a meta-RL algorithm.
% , namely Domain Randomization and Meta Reinforcement Learning. 

\subsection{Domain Randomization}
According to \cite{Zhao2020}, DR has emerged as the most prevalent technique to bridge the reality gap in the field of robotics. 
Given the lack of literature focusing specifically on the reality gap within the TSC domain, it is an intuitive step to start the investigation with DR as a well-established method from another domain.
% Since there has been little literature on the reality gap and methods to overcome it in the TSC domain, DR therefore represents a promising lead. 
To our knowledge, DR for TSC has only been used in \cite{Garg2022}, where the focus was on the visual domain, similar to \cite{Tobin2017}. No consideration of traffic dynamics or varying traffic volumes was undertaken. Similarly, no state information such as the waiting time of vehicles was included either, which cannot be extracted from images alone. Therefore, the findings from \cite{Garg2022} are not applicable to our investigation.

DR is a technique in reinforcement learning where an agent is exposed to various environments that use randomized simulator's parameters (e.g. car-following models, traffic volume, or sensor noise). 
% The motivation for DR comes from the fact that the exact parameters of the reality (target domain) are unknown anyway. 
Since the exact parameters of the reality (target domain) are unknown anyway, the agent should learn to perform well over a set of randomized environments in simulation (source domain), so that it will be able to generalize well in a real-world scenario. The target domain can be seen as one instance of the randomized environments \cite{Muratore2021}\cite{Tobin2017}. The fact, that DR methods can be deployed at the target domain without any further adaptation step constitutes them a so-called \textit{zero-shot learning} method. 

% TODO: Make mathematical notation consinsten with RL chapter
For the mathematical description of DR, we define a distribution over the environment parameters, denoted by $M$. Each parameter setting $\mu \in M$ defines a specific instance of the environment respectively MDP $(S_\mu, A_\mu, P_\mu, R_\mu, \gamma_\mu)$. The objective of DR is to learn a policy $\pi$ that performs well across the distribution of environments rather than just in a single environment. Thus, the optimization problem can be written as:
\begin{equation}    
\pi^* = \arg\max_{\pi} \mathbb{E}_{\mu \sim M} \left[ \mathbb{E}_{\tau \sim \pi(\cdot|\mu)} \left[ \sum_{t=0}^{T} \gamma_\mu^t R_\mu(s_t, a_t) \right] \right],
\end{equation}

where $\tau = (s_0, a_0, s_1, a_1, \ldots, s_T, a_T)$ is a trajectory generated by the policy $\pi$ in an environment with parameter setting $\mu$, and $T$ is the time horizon of the RL problem.

\subsection{Meta Reinforcement Learning}
Meta-RL has generated a lot of research interest in recent years, and overcoming the reality gap is only one of the possible application areas. It is a subfield of reinforcement learning focused on developing algorithms that can efficiently adapt to new tasks by leveraging prior experience. In \cite{Arndt2020} for example, a robotics policy trained in simulation was transferred to real hardware with a domain adaption technique based on Meta-RL. 
In the context of RL-based TSC, Meta-RL was investigated to increase the learning speed of RL by utilizing previously acquired knowledge \cite{Zang2020}. In addition, \cite{Kim2022} investigated the use of Meta-RL for recognizing the current traffic situation and adjusting the reward function accordingly. The usage of Meta-RL to overcome the reality gap in the field of RL-based TSC was not investigated yet.

Meta-learning aims to train a model on a range of learning tasks, enabling it to tackle new learning tasks by utilizing only a limited amount of training samples \cite{Finn2017}. In the context of bridging the reality gap, the source domain comprises randomized environments, each treated as a learning task. The agent trained on different learning tasks is expected to learn the target domain, which represents reality in our case, through a few so-called knowledge updates. These updates adapt the policy to the dynamics of the real-world environment. Unlike DR, Meta-RL is therefore categorized as a \textit{few-shot learning} method.

In our study, we employ MAML as our Meta-RL algorithm \cite{Finn2017}. MAML is considered one of the most influential and widely adopted Meta-RL algorithms \cite{Ye2022}. Its popularity is evidenced by the development of numerous extensions and follow-up works \cite{Beck2023}, and its incorporation into well-known reinforcement learning libraries such as RLlib \cite{Liang2018}, and other machine learning libraries \cite{higher2019}\cite{Arnold2019}\cite{garage2019}.
% such as higher \cite{higher2019}, learn2learn \cite{Arnold2019}, and garage \cite{garage2019}. 
% MAML's versatility contributes to its widespread use. As a model-agnostic algorithm, MAML can be applied to any model trainable with gradient-based methods, making it suitable for various applications across reinforcement learning, supervised learning, and unsupervised learning \cite{Beck2023}.

MAML is designed to train an agent's initial parameters in a way that enables it to achieve good generalization performance on a new task after only a few gradient steps with some data from that task \cite{Finn2017}. 
% The core concept of this approach is to explicitly train the model's parameters so that they can be easily fine-tuned, ultimately leading to maximal performance on new tasks after updating the parameters through one or more gradient steps with limited data from the new task \cite{Finn2017}.
MAML's mathematical formulation consists of two nested optimization loops. In the inner loop, the parameters $\theta$ are updated for each task $i$ using the task-specific loss function $L_i$ and a small amount of training data: 
\begin{equation}
\theta_i' = \theta - \alpha \nabla_\theta L_i(\theta),
\end{equation}
where $\alpha$ represents the inner loop learning rate. In the outer loop, the objective is to minimize the average loss across all tasks after the inner loop updates: 
\begin{equation}
\theta^* = \arg\min_\theta \sum_{i=1}^N L_i(\theta_i'),
\end{equation}
where $N$ denotes the number of tasks.  
This is also done by using gradient-based optimization methods. The learning rate of the outer loop is denoted by $\beta$.

\section{Experiments}
\label{sec:experiments}
The primary goal of our experiments is to assess the effectiveness of DR and MAML in bridging the reality gap in the TSC domain. To this end, we employ the SUMO model of an intersection in Lemgo, integrated within the LemgoRL framework, as our source domain. We train both DR and MAML using this model, with the aim of optimally controlling the traffic at the intersection. 
As the base algorithm for both methods, we use PPO, which is a state-of-the-art and widespread algorithm in the RL community.

% To establish a baseline for comparing the performance of DR and MAML, we train also a PPO agent without any randomization of the parameters in the source domain. 

During the training process, we randomize a variety of parameters 
% that influence driving behaviors, vehicle properties, sensor noise, and traffic demand, as 
described in section \ref{sec:RealityGap}. Once trained, the agents are deployed and evaluated within the target domain, which is a simulation model of the same intersection created using the professional traffic simulation tool VISSIM \cite{Vissim2010}. Therefore, we integrate the VISSIM model of the intersection into the LemgoRL framework.
% , replacing the SUMO model.

We chose VISSIM to create a model as our target domain for several reasons. First, it is considered one of the most accurate and realistic traffic simulation tools available. This means that if the methods can successfully transfer a policy trained in SUMO to VISSIM, it is highly likely that they would perform similarly in a real-world setting. Second, to quantify the effectiveness of the methods, it is not practical to deploy the policies in real-world traffic scenarios, where non-reproducible traffic volumes occur. This would make a quantifiable comparison between the methods less precise.

An ideal baseline for performance comparison of DR and MAML would be to train a PPO agent directly in VISSIM. However, due to certain technical constraints, training with VISSIM is enormously time-consuming. For instance, the parallelization of multiple instances is highly limited. Also, it is not possible to run the simulation in non-GUI mode. Given these limitations, we had to refrain from training directly in VISSIM. Consequently, for the baseline, we resort to a PPO agent trained in SUMO without any measures to reduce the reality gap.

\begin{figure}
    \centering
    \includegraphics[width=0.49\textwidth]{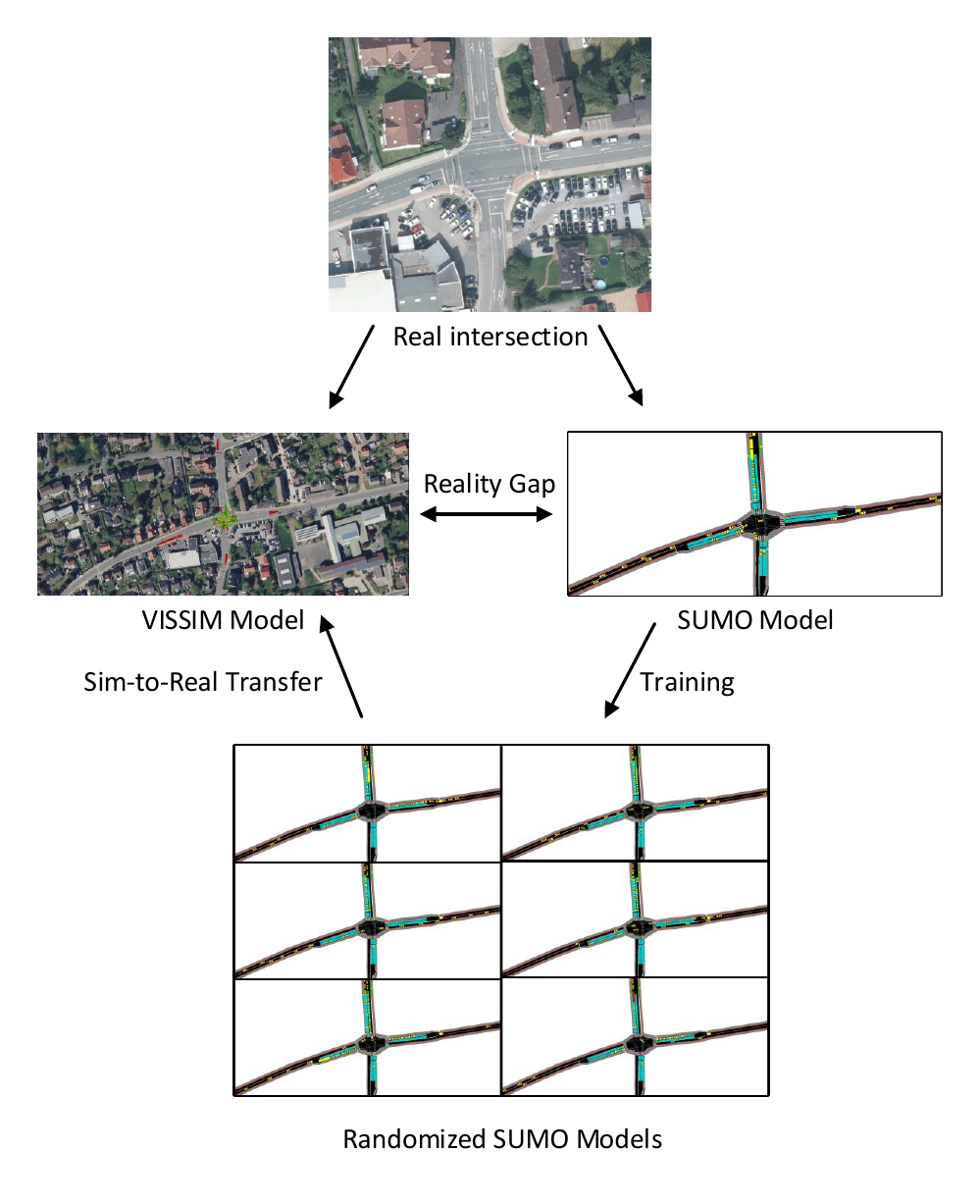}
    \caption{Conceptual overview of the experiment design}
    \label{fig:conceptual_view}
\end{figure}

See Fig.~\ref{fig:conceptual_view} for a conceptual overview of the experiment design.

\subsection{Parameter Randomization}
% During the training phase of DR and MAML, specific parameters that affect driving behavior, vehicle characteristics, and traffic volume in the simulation are randomly varied. 
We define different distributions to capture the variability of the considered simulation parameters.
The parameters involved, along with their respective ranges and distributions, are detailed in Table~\ref{tab:random_parameters}.
We use the notation $U_C(a, b)$ to denote a uniform distribution for continuous values between $a$ and $b$. For sampling the car-following model, we use a discrete uniform distribution denoted with $U_D({\cdot})$. 
% Some of these parameters are sampled uniformly from continuous or discrete ranges of values. , symbolized by $U$ and $Uniform$, respectively. 
Other parameters follow a normal distribution, denoted by $\mathcal{N}(mean, var)$ with specified mean and variance. 
If the sampled values are restricted to a specific interval $[min, max]$, we utilize the function $clip(\cdot, min, max)$.
Note that certain parameters, such as \texttt{length}, which are related to vehicle classes, require individual distributions for each class. In table~\ref{tab:random_parameters}, only the values for the vehicle class \texttt{car} are shown.

Every real sensor system exhibits noise \cite{Dulac-Arnold2021}. To ensure that agents in real systems can deal with this robustly, noise is typically added to sensor signals during training \cite{Dulac-Arnold2021}\cite{Peng2018}\cite{Tan2020}. For our model, we assume the usage of camera-based sensors. In SUMO, these are modeled by \texttt{laneAreaDetectors}. 
% The following noise mechanisms are introduced during training to represent the versatile occurrence of noise in camera-based sensors. 
We implemented the following noise mechanisms:

\begin{enumerate}
    \item Each waiting and non-waiting vehicle is missed with a probability of $\eta_{queue}$ and $\eta_{wave}$, respectively, which in turn affects the values of \texttt{queue} and \texttt{wave}. This corresponds to non-detection of vehicles due to bad weather conditions or similar.
    \item For the \texttt{laneAreaDetectors} the parameter \texttt{speedThreshold} can be used to set the speed value from which a vehicle should be classified as "queued". However, due to inaccuracies, \texttt{speedThreshold} may be slightly shifted in real sensor systems. Therefore we randomize this value.
    \item Moreover, at each time step, with a probability of $\eta_{queue,pos}$, a non-waiting vehicle is misclassified as waiting, so that \texttt{queue} is increased. 
    \item Determining the waiting time of vehicles \texttt{wait\_veh} is typically implemented using object tracking algorithms that track vehicles across multiple frames. Intermediate false detections of vehicles may result in not detecting the full duration of the waiting time. Therefore, at each step we reduce the waiting time by a small duration sampled from $U_D({2,6,10})$s. This reduction is done with a probability of $\eta_{wait\_veh}$.
    \item When pedestrians want to cross the intersection, they typically press a demand button. This operation is usually not noisy. Therefore, we do not add noise to \texttt{wait\_ped}.
\end{enumerate}
For each episode, new values are sampled for these noise-related parameters (see Table~\ref{tab:random_parameters}):

\begin{table}[ht]
\centering
\caption{Randomized simulation parameters and their respective ranges. Vehicle-related parameters are only shown for the vehicle class \texttt{car}.}
\begin{tabular}{@{}ll@{}}
\label{tab:random_parameters}
% \toprule
\textbf{Parameter}  & \textbf{Range and Distribution}\\
\hline
Car-Following Model & $U_D(\{\text{Krauss}, \text{IDM}\})$ \\
\texttt{tau} & $\text{clip}(\mathcal{N}(1.13, 0.1^2), 1.0, 1.2)$ \\ 
\texttt{sigma} (Krauss only) & $\text{clip}(\mathcal{N}(0.59, 0.23^2), 0.0, 1.0)$ \\
\texttt{delta} (IDM only) & $\text{clip}(\mathcal{N}(3.97, 0.05^2), 3.7, 4.3)$ \\
\texttt{speedFactor} & $\text{clip}(\mathcal{N}(1.06, 0.11^2), 0.6, 1.4)$ \\
\texttt{minGap} & $\text{clip}(\mathcal{N}(2.9, 0.36^2), 1.5, 4)$ \\
\texttt{jmStoplineGap} & $\text{clip}(\mathcal{N}(0.94, 0.26^2), 0.5, 2)$ \\
\texttt{impatience} & $\text{clip}(\mathcal{N}(0.29, 0.15^2), -0.1, 0.5)$ \\
\texttt{accel}  & $\mathcal{N}(2.5, 0.21^2)$ \\
\texttt{decel} & $\mathcal{N}(4.7, 0.21^2)$ \\
\texttt{length} & $\text{clip}(\mathcal{N}(5.0, 0.64^2), 4.7, 5.0)$ \\
\texttt{scale} & $U_C(0.85, 1.18)$ \\
\texttt{speedThreshold} & $U_C(3/3.6, 10/3.6)$ \\
$\eta_{queue}$ & $U_C(0, 0.06)$ \\
$\eta_{queue,pos}$ & $U_C(0, 0.06)$ \\
$\eta_{wave}$ & $U_C(0, 0.06)$ \\
$\eta_{wait\_veh}$ & $U_C(0, 0.1)$ \\
\end{tabular}
\end{table}

% \begin{figure}
%     \centering
%     \includegraphics[width=0.48\textwidth]{graphics/randomization_types.jpg}
%     \caption{Different Randomization Types.}
%     \label{fig:random_types}
% \end{figure}

\subsection{Training Protocol and Network Architecture}
% - In order to integrate DR into PPO, the PPO algorithm is used to optimize the policy over the distribution of environments instead of a single environment. At each iteration, a batch of environment instances is sampled from the parameter distribution $\Theta$. 

For the RL training, each episode spans over a simulation duration of 3600 seconds (corresponding to one hour of real-time), with each timestep equal to one second. PPO and DR are trained with 10,000 episodes, tallying up to 36 million timesteps in total. This total number of timesteps is used for the training of MAML as well, to ensure fair comparison between all approaches. At the onset of each episode, a new set of parameters $\mu$ is sampled for DR or a new task with pre-sampled parameters is selected for MAML.

For the hyperparameters used in this study, we mostly followed \cite{Mueller2022}. The neural network is configured with two hidden layers, each consisting of $128$ neurons. The activation function employed throughout the network is the hyperbolic tangent (tanh). The training phase utilizes a batch size of $2048$, with mini-batches of $512$, and an ADAM optimizer \cite{Kingma2014} with a learning rate of $4e-5$ is used to update the network parameters. 
% For the most part, we have followed \cite{Mueller2022} in the choice of these hyperparameters. 

% Absatz "Trainingsprotokoll" ähnlich wie bei Peng2018 formulieren: "At the start of each episode, a new set of parameters μ is sampled by drawing values for each parameter from their respective range."
% - Trainingsprotokoll:
%     - episode length of 3600s, 1 timestep = 1s, 3600 Simulationsdauer entspricht also einer Stunde
%     - number of episodes for run with PPO=10000 and run with DR=10000
%     - accordingly MAML = 36 Mio timesteps, entspricht 3600 * 10000
%     - Zu Beginng jeder Episode wird ein neuer Parametersatz µ gesampelt (DR) bzw. eine neue zufällige Task-ID mir vorher zufällig gesampelten Parameters ausgewählt (Meta-RL)

% Absatz "Aufbau des Netzes" ähnlich wie bei Peng2018 formulieren: "During training, parameter updates are performed using the ADAM optimizer [40] with a stepsize of5×10−4 for both the policy and value function. Updates are performed using batches of 128 episodes with 100 steps per episode."
% - Aufbau des Netzes:
%     Training batch size 2048
%     mini batch size	512
%     learning rate 0.00004
%     activation function tanh
%     2 hidden layers with 128 neurons each

\subsection{Evaluation Settings}
In the evaluation phase, the trained agents are subjected to the target domain. One difference between the source and target domain relates to the intersection geometry. In SUMO, the model is based on an import from OpenStreetMap, while in VISSIM, the model is manually created based on satellite images, leading to a more accurate representation. It is worth noting that the intersection geometry was not randomized during training. Another difference comes from the driving behavior. The car-following model used in the evaluation phase is the one developed by Wiedemann \cite{Wiedemann1974}, which is the standard model in VISSIM, while the source domain uses the Krauss model and IDM. For all other parameters relating to driving behavior and vehicle properties, we use the default VISSIM parameters. The traffic volume used during evaluation corresponds to $\texttt{scale}=1$. 

% \textbf{Traffic Volume:} The same traffic volume is modeled as in the SUMO simulation with $\texttt{scale}=1$.

% \textbf{Intersection Geometry:} The intersection geometry differs between the training and evaluation settings due to the use of different simulation environments. In SUMO, the model is based on an import from OpenStreetMap (as detailed in the Reality Gap chapter), while in VISSIM, the model is manually created based on satellite images, leading to a more accurate representation. It's worth noting that the intersection geometry was not randomized during training.

% \textbf{Driving Behavior:} The car-following model used in the evaluation phase is the one developed by Wiedemann \cite{Wiedemann1974}, which is the standard model in VISSIM.

% \textbf{Vehicle Properties:} The default settings of VISSIM are used for vehicle properties.

For each evaluation run, the traffic volume, vehicle properties, driving behavior, and intersection geometry remain identical. 
% We investigate the same VISSIM model with three different sensor noise randomizations ($\eta_{queue}$, $\eta_{queue,pos}$, $\eta_{wave}$, $\eta_{wait\_veh}$, and \texttt{speedThreshold}). 
We investigate the same VISSIM model with three different sensor noise settings sampled from $\eta_{queue}$, $\eta_{queue,pos}$, $\eta_{wave}$, $\eta_{wait\_veh}$, and \texttt{speedThreshold}.  
The specific values for these parameters were withheld during training so that the algorithms could not adjust to them.
% , as was done in \cite{Dai2019}. 
In the evaluation phase, therefore, the same challenge exists as in reality, since the exact noise parameters are not known here either. We refer to the different values for sensor noise as evaluation setting A, B, and C in the following.
The evaluation is conducted over five episodes with different random seeds, with each agent experiencing the same set of random seeds. Each episode lasts for 3600 seconds, equivalent to an hour.

\subsection{Results and Discussion}
Refer to Table \ref{tab:metrics} for a detailed comparison of the performance of the different algorithms across all evaluation settings.

\begin{table}[!ht]
    \centering
    \caption{Evaluation results: The average, taken over all lanes and pedestrian crosswalks and across 5 simulation runs, is shown for each evaluation setting and algorithm. MAML ft. represents the fine-tuned version of MAML.}
    \label{tab:metrics} 
    \begin{tabular}{@{}ccccccc@{}}
    \toprule
    \thead{Evaluation \\ Setting} & \thead{Algorithm} & \thead{$\texttt{queue}$ \\ ${[m]}$} & \thead{$\texttt{wait}$ \\ $\texttt{\_veh} {[s]}$} & \thead{$\texttt{wait}$ \\ $\texttt{\_ped} {[s]}$} & \thead{Cum. \\ Reward}  \\ \midrule
    \multirow{4}{*}{A} & PPO & 14.1 & 215.3 & 3.7 & -1222.1  \\ 
    & DR & 10.3 & 138.2 & 3.3 & -846.0 \\ 
    & MAML & 9.3 & 93.8 & 5.1 & -686.3 \\ 
    & MAML ft. & 10.3 & 112.9 & 4.5 & -781.2 \\ \midrule
    \multirow{4}{*}{B} & PPO & 13.6 & 209.5 & 3.4 & -1186.1 \\ 
    & DR & 9.6 & 130.2 & 3.5 & -793.6 \\  
    & MAML & 9.4 & 103.3 & 5.1 & -719.2 \\ 
    & MAML ft. & 9.4 & 99.9 & 4.5 & -708.1 \\ \midrule
    \multirow{4}{*}{C} & PPO & 13.3 & 193.7 & 3.4 & -1130.3 \\ 
    & DR & 10.2 & 141.5 & 3.4 & -848.3 \\  
    & MAML & 9.4 & 97.8 & 4.6 & -702.3 \\ 
    & MAML ft. & 8.9 & 90.3 & 5.0 & -657.5 \\ \midrule
    \multirow{4}{*}{Avg.} & PPO & 13.7 & 206.2 & 3.5 & -1179.5 \\ 
    & DR & 10.0 & 136.6 & 3.4 & -829.3 \\  
    & MAML & 9.4 & 98.3 & 4.9 & -702.6 \\ 
    & MAML ft. & 9.5 & 101.0 & 4.7 & -715.6 \\ \midrule
    \multirow{4}{*}{\thead{Avg. \\ Relative}} & PPO & 100.0\% & 100.0\% & 100.0\% & 100.0\% \\ 
    & DR & 73.4\% & 66.3\% & 97.5\% & 70.3\% \\  
    & MAML & 68.6\% & 47.7\% & 142\% & 59.6\% \\  
    & MAML ft. & 69.7\% & 49.0\% & 135.1\% & 60.7\% \\  
    \bottomrule
\end{tabular}
\end{table}

We first compare PPO and DR. On average, DR outperforms PPO in all metrics. While the improvement in \texttt{wait\_ped} is marginal, significant improvements are noted in vehicle-related metrics. The avg. value for \texttt{queue} in DR is 26.6\% lower than in PPO, for \texttt{wait\_veh} even 33.7\% lower. This means that DR significantly improves the metrics for motorized traffic, while the waiting time for pedestrians remains similar.

Moving to the results obtained by MAML, we encountered that successfully training this algorithm for 3600 seconds was not feasible despite extensive hyperparameter tuning. The agent failed to learn. Only upon reducing the episode duration to 1800 seconds, the algorithm started to show learning capabilities. 
The reason for MAML's sensitivity to episode duration should be investigated in further research. A possible remedy could be to add a few linear layers to the neural network, because according to \cite{Arnold2021Maml} this can help to support the meta-learning process. 
To expose MAML to all traffic patterns included in the 3600s training duration despite the shortened episode duration of 1800s, the start time is randomized at the beginning of each episode. Thus, for each episode, MAML goes through a different time period and experiences all traffic patterns over the course of the training.

We also found that MAML is highly sensitive to the hyperparameters \texttt{inner\_adaption\_steps} and \texttt{maml\_optimizer\_steps}, which refer to the number of iterations in the inner loop and outer loop respectively. The combination we found with the highest cumulative reward during training was $\texttt{inner\_adaption\_steps}=2$ and $\texttt{maml\_optimizer\_steps}=20$.

As MAML is a few-shot learning algorithm, we fine-tuned the agents for an additional five episodes in the specific evaluation setting. This seems a realistic magnitude for fine-tuning if such systems are to be deployed in real-world scenarios. A larger number of fine-tuning episodes would require accepting a potentially longer time of deploying a suboptimal TSC policy.

% One notable issue is that training happens for 1800 seconds, but evaluation is carried out over 3600 seconds. The reason we insist on 3600 seconds is to train agents for a period where the traffic dynamics are quite similar, for instance, during rush hour. Training focused on shorter durations does not make sense as such short periods do not exhibit specific characteristics that are reproducible in reality. Instead, it is more important to learn the basic characteristic of a traffic pattern (variations within a rush hour) rather than focusing too specifically on a short time.

\begin{figure}
    \centering
    \includegraphics[width=0.49\textwidth]{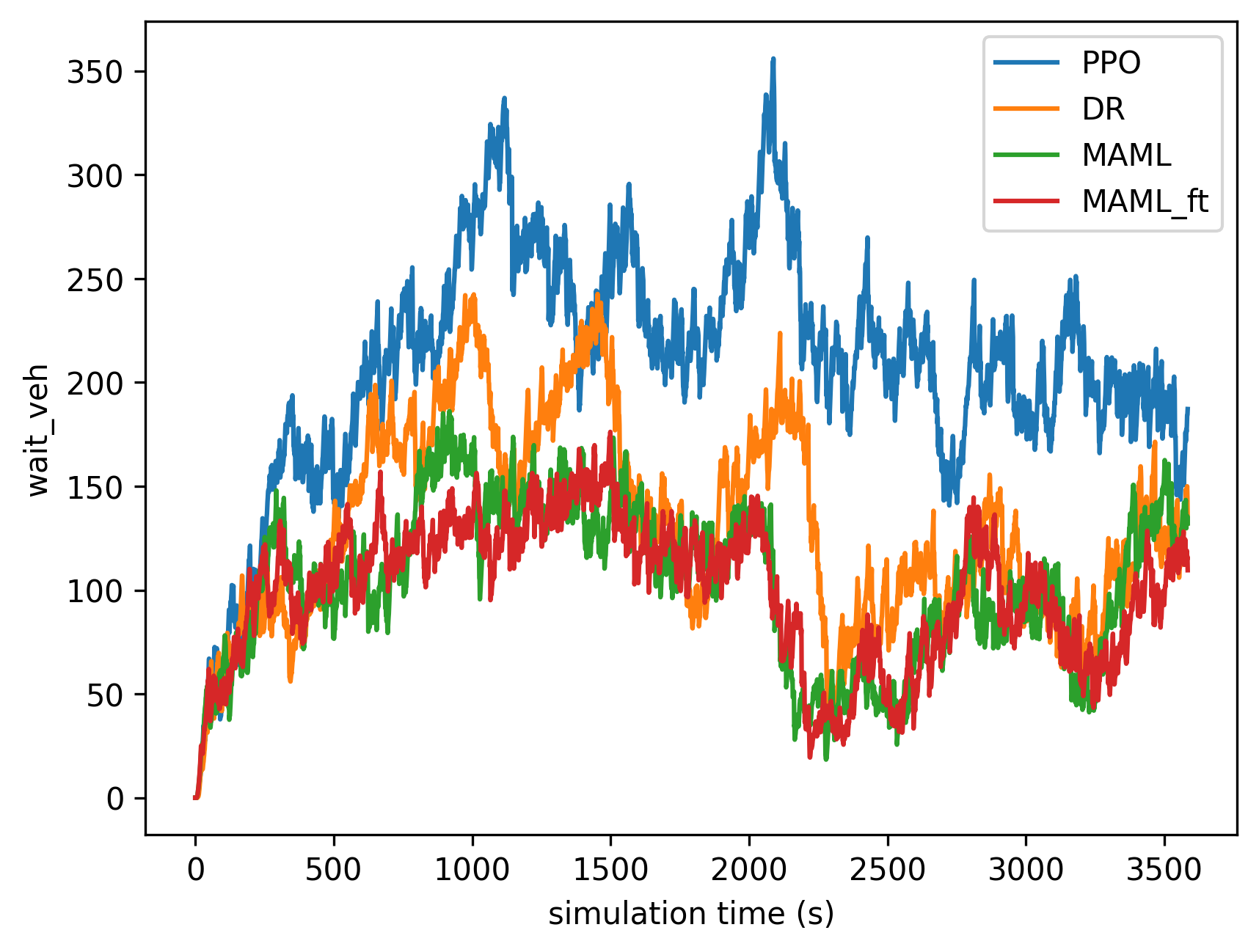}
    \caption{\texttt{wait\_veh} for all algorithms in evaluation setting B. The curves show the average of 5 episodes.}
    \label{fig:waitveh}
\end{figure}

Examining the MAML performance, it can be said that both the baseline and the fine-tuned version perform better in terms of cumulative reward than PPO and DR. This is achieved by MAML accepting longer wait times for pedestrians - up to 42\% - and thereby prioritizing vehicles, which results in smaller values for \texttt{queue} (31.4\% less than PPO) and \texttt{wait\_veh} (52.3\% less than PPO). 
% In fig.~\ref{fig:waitveh}, \texttt{wait\_veh} for all algorithms in evaluation setting B is depicted over the simulation time.  
Fig.~\ref{fig:waitveh} shows the course of \texttt{wait\_veh} for all algorithms in the evaluation setting B.  
% It is noteworthy, that MAML outperforms DR and PPO although it is trained only on 1800s.

Interestingly, fine-tuning MAML only brought further improvement over the base version in setting B and C. In setting A, fine-tuning led to better \texttt{wait\_ped} but worse vehicle-related metrics, resulting in a worse cumulative reward. This could be due to a fundamental issue common to all methods requiring fine-tuning: The reward function consists of information that can only be obtained through sensors. In training, we have access to the true sensor values, which we use for calculating the reward. But, during fine-tuning, it happens in a noisy environment where the agent has no access to the true values. This means that fine-tuning is conducted with noisy reward values. Since MAML is trained to be sensitive to gradient updates during fine-tuning, a few updates with noisy reward can degrade the performance of the algorithm.

\section{Conclusion}
\label{sec:conclusion}
This paper represents the first study investigating the reality gap for RL-based TSC. It includes a broad spectrum of parameters that may influence this gap and explored two widely used methods for bridging the reality gap, known in particular from robotics: Domain Randomization and Model-Agnostic Meta-Learning. The ability of these methods to bridge the reality gap in the TSC domain was examined using LemgoRL, which to our knowledge, represents the most mature RL framework in terms of its real-world applicability. The target domain is a model implemented in VISSIM replicating reality. Our experiments demonstrated that both DR and MAML significantly outperform the standard PPO algorithm, proving their suitability for bridging the reality gap.

In terms of practical implementation and training, DR is an easier approach to use, even though its performance did not match that of MAML. On the other hand, despite its better results, MAML encountered its limitations when training for longer episodes and was also sensitive to certain hyperparameters. An additional challenge is fine-tuning MAML with noisy rewards. Therefore, further research is needed to improve MAML to take advantage of its superior performance while overcoming these identified challenges.
% - DR ist einfacher in der Handhabung, erzielt aber nicht so gut Ergebnisse wie MAML
% - MAML hingegen konnte nicht für lange Episoden trainiert werden und hat sich zudem als recht Hyperparameter sensitiv erwiesen
% - Hinzu kommt, das Problem mit fine-tuning in einer noisy Umgebung
% - Daher ist weitere Forschung notwendig, um MAML so zu erweitern, dass die bessere Performance von MAML genutzt werden, gleichzeitig aber die genannten Herausforderungen adressiert werden

Another emergent research question from our work is the identification of key simulation parameters that should be encompassed in the randomization process. We hope our paper will stimulate further research to advance RL-based TSC towards real-world applications.

\section*{ACKNOWLEDGMENT}
We would like to thank Vishal Rangras and Nehal Soni who supported us with their expertise and experience.

%%%%%%%%%%%%%%%%%%%%%%%%%%%%%%%%%%%%%%%%%%%%%%%%%%%%%%%%%%%%%%%%%%%%%%%%%%%%%%%%
% \bibliographystyle{IEEEtran111.bst}
\bibliographystyle{IEEEtran} % if the IEEEtran.bst package provided by the conference site should be used(v1.11), then upload it and specify the parameter with IEEEtran.bst. Otherwise the newest version (V1.14) is automatically chosen by overleaf.
\balance
% \bibliography{references.bib}
\bibliography{references}
% \begin{thebibliography}{99} % not recommended since I have to insert the bib items by my own
% \end{thebibliography}

\end{document}